\documentclass[review]{elsarticle}
\usepackage{lineno,hyperref}
\usepackage{graphicx}
\usepackage{subfigure}
\usepackage{multirow}
\usepackage{tabu}

\usepackage{graphicx}
\usepackage{comment}
\usepackage{amsmath,amssymb} 
\usepackage{color}

\modulolinenumbers[5]

\journal{}

\begin{document}

\begin{frontmatter}

\title{
	Leveraging Local and Global Descriptors in Parallel to Search Correspondences for Visual Localization
}




\author[mymainaddress,mysecondaryaddress]{Pengju Zhang}

\author[mymainaddress,mysecondaryaddress]{Yihong Wu\corref{mycorrespondingauthor}}
\cortext[mycorrespondingauthor]{Corresponding author}
\ead{yhwu@nlpr.ia.ac.cn}

\author[mymainaddress,mysecondaryaddress]{Bingxi Liu}

\address[mymainaddress]{School of Artificial Intelligence, University of Chinese Academy of Sciences, Beijing 100049, China}
\address[mysecondaryaddress]{National Laboratory of Pattern Recognition, Institute of Automation, Chinese Academy of Sciences, Beijing 100190, China}

\begin{abstract}
	
Visual localization to compute 6DoF camera pose from a given image has wide applications such as in robotics, virtual reality, augmented reality, etc. Two kinds of descriptors are important for the visual localization. One is global descriptors that extract the whole feature from each image. The other is local descriptors that extract the local feature from each image patch usually enclosing a key point. More and more methods of the visual localization have two stages: at first to perform image retrieval by global descriptors and then from the retrieval feedback to make 2D-3D point correspondences by local descriptors. The two stages are in serial for most of the methods. This simple combination has not achieved superiority of fusing local and global descriptors. The 3D points obtained from the retrieval feedback are as the nearest neighbor candidates of the 2D image points only by global descriptors. Each of the 2D image points is also called a query local feature when performing the 2D-3D point correspondences. In this paper, we propose a novel parallel search framework, which leverages advantages of both local and global descriptors to get nearest neighbor candidates of a query local feature. Specifically, besides using deep learning based global descriptors, we also utilize local descriptors to construct random tree structures for obtaining nearest neighbor candidates of the query local feature. We propose a new probabilistic model and a new deep learning based local descriptor when constructing the random trees. A weighted Hamming regularization term to keep discriminativeness after binarization is given in the loss function for the proposed local descriptor. The loss function co-trains both real and binary descriptors of which the results are integrated into the random trees. Experiments on challenging benchmarks show that the proposed localization method can significantly improve the robustness and accuracy compared with the ones which get nearest neighbor candidates of a query local feature just based on either local or global descriptors.

\end{abstract}


\begin{keyword}
	\texttt{Visual localization}\sep \texttt{6DoF pose}\sep \texttt{Parallel search} \sep \texttt{Learning based descriptor}
\end{keyword}

\end{frontmatter}

%

\section{Introduction}
Visual localization, estimating 6DoF camera pose from an image in a known 3D scene, has wide applications, such as in augmented reality \cite{middelberg2014scalable}, virtual reality, automated navigation \cite{munoz2020ucoslam}, etc. 
Visual localization methods can be divided into: end-to-end learning ones and non-end-to-end ones. The non-end-to-end ones include indirect image retrieval based methods (abbreviated to indirect), direct 2D-3D correspondence methods (abbreviated to direct), and indirect-direct fusion methods \cite{piasco2018survey}. Usually, the end-to-end learning methods and the indirect methods use global descriptors that extract the whole image features. The direct methods usually use local descriptors that extract the local features from image patches enclosing key points. The given image to localize its camera is also called a query image especially when performing image retrieval. When performing 2D-3D correspondences, each point in the given image is also called a query point and its local feature is also called a query local feature.

The end-to-end learning methods \cite{kendall2015posenet} aim to straightforward regress the camera pose of a query image by training convolutional neural networks (CNNs) with a set of training images and their corresponding poses. The indirect methods \cite{arandjelovic2016netvlad, torii201524} first retrieve images \cite{celik2017content}, then the location of the query image is estimated from the geo information of the most relevant retrieved image. The direct methods \cite{sattler2016efficient,svarm2016city} establish correspondences between 2D points in a given query image and 3D points in databases which have been reconstructed offline, then the 6DoF camera pose of the query image is computed by RANSAC and PnP algorithms. 

\begin{figure}
	\centering
	\subfigure{
		\label{Fig.sub.1}
		\includegraphics[width=0.4\textwidth]{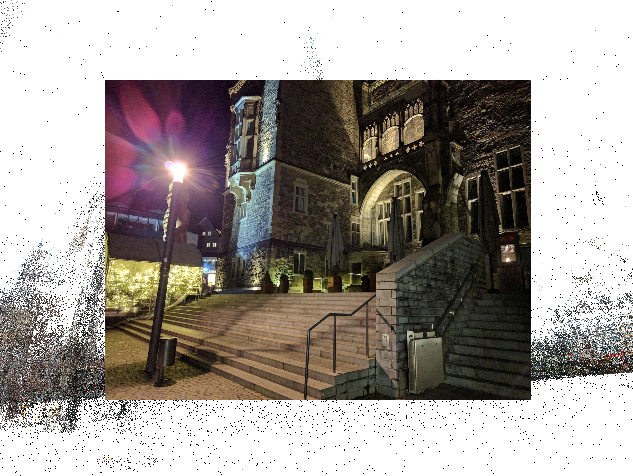}}
	\subfigure{
		\label{Fig.sub.2}
		\includegraphics[width=0.4\textwidth]{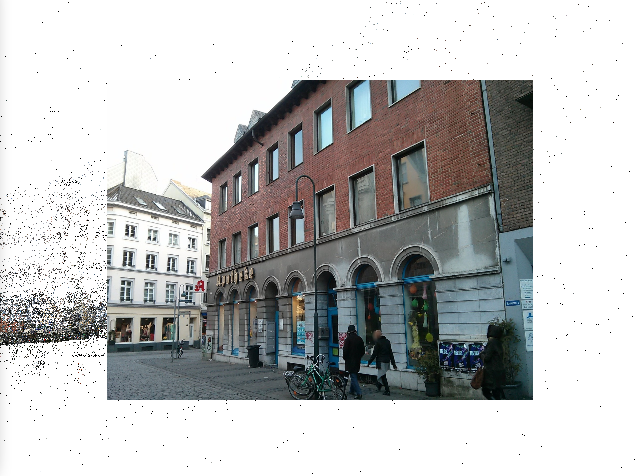}}
	
	\subfigure{
		\label{Fig.sub.3}
		\includegraphics[width=0.4\textwidth]{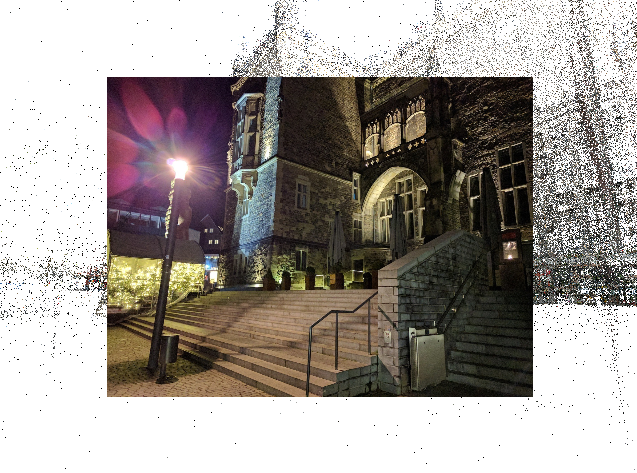}}
	\subfigure{
		\label{Fig.sub.4}
		\includegraphics[width=0.4\textwidth]{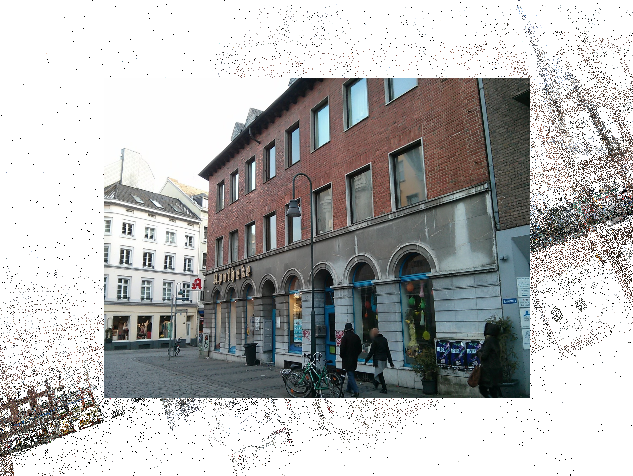}}
	\caption{Two query images are aligned with 3D points according to the poses estimated by a direct method (RT\_AP+CRBNet, a random tree structure to search 2D-3D correspondences) and a serial indirect-direct method (IR+CRBNet). The first line shows the localized results obtained by the direct method which succeeds for a viewpoint change but fails for an illumination change. The second line shows the localized results obtained by the indirect-direct method which succeeds for the illumination change but fails for the viewpoint change. The images are taken from the Aachen Day-Night dataset \cite{sattler2018benchmarking}.(Best viewed in color)}
	\label{motivation}
\end{figure}

By adding some images taken in nighttime into training process, the end-to-end methods and the indirect methods are robust to illumination changes. But due to these methods are highly dependent on global descriptors, they may be not robust to large viewpoint changes. The direct methods, highly dependent on local descriptors when selecting nearest neighbor candidates of a given query local feature, are accurate on large viewpoint changes but not robust to illumination or season changes. This is because local descriptors contain relatively little information making sensitive to illumination changes. It follows that indirect-direct methods fusing the indirect and direct methods are appearing more and more. Most of these methods are in the following two stages \cite{sarlin2019coarse}:  first performing image retrieval from databases by learning based global descriptors and then performing 2D-3D correspondences from the retrieval feedback by learning based local descriptors. The two stages are in serial and the 3D points obtained from the  retrieval feedback as the nearest neighbor candidates of the 2D query points are only gotten from global descriptors. This simple combination of the indirect and direct methods has not achieved superiority of fusing local and global features. For example, the range of nearest neighbor candidates of a query local feature is a key part but is scoped by the first stage. When the viewpoint of the query image varies greatly from the database images, the serial indirect-direct methods are likely to fail because of the incorrect retrieved images. Fig. \ref{motivation} shows two examples of localization results of a serial indirect-direct method and a direct method, where the indirect-direct method succeeds for the illumination change but fails for the viewpoint change and the direct method succeeds for a viewpoint change but fails for an illumination change. The used direct method is based on Feng et al. \cite{feng2015fast}. It constructs random trees by supervised information of binary local descriptors to cluster 3D points into different leaf nodes and gives a fast search of 2D-3D correspondences.

By fully considering all the above problems and motivated by the random tree method of Feng et al. \cite{feng2015fast}, we propose an efficient visual localization method to compute the 6DoF camera pose from a given query image. Global descriptors of the query image and images from database are all extracted. Local descriptors of key points in the given query image are also extracted. We fuse indirect and direct methods in parallel, which is different from the previous methods fusing in serial. We search nearest neighbor candidates of a given query local feature by both local and global descriptors in parallel. Local descriptors in databases are used to construct random trees and learning based global descriptors for images are used to perform image retrieval. Then prior leaf nodes in the random trees, which most likely contain the nearest neighbors of the local query feature, are obtained by a probability model. Simultaneously, prior frames, i.e. K-Nearest Neighbor (KNNs) in database of the query image are obtained by learning based global descriptors.
Subsequently, 3D points in the prior leaf nodes and 3D points visible by the prior frames make up the whole nearest neighbor candidates of the query local feature. 
The way of random trees searches nearest neighbor candidates of the given query local feature by local descriptors while the way of image retrieval searches nearest neighbor candidates of a given query local feature by global descriptors. These two ways can complement each other and their fusion can avoid missing the true correspondences.

We also propose a weighted Hamming regularization term to keep discriminativeness after binarization for improving a learning local descriptor. The local descriptor can Co-train Real-valued and Binary forms together for image keypoints. We name it CRBNet. The binary CRBNet is used to construct the random tree structure. Based on CRBNet, a new probabilistic model is given to calculate which leaf nodes of the random trees have the best probability to contain the nearest neighbor of a query local feature. The real-valued CRBNet is used to compute the Euclidean distance between the query local feature and its nearest neighbor candidates. 

The new contributions of this paper are summarized as: 1) We present a novel visual localization framework which fuses local and global descriptors in parallel to find nearest neighbor candidates of a given query local feature.
2) We propose a weighted hamming regularization term when training learning based local descriptors in order to keep discriminativeness after binarization. And then a CRBNet that co-trains real-valued and binary descriptors together is given. 
3) A new probability model is established for searching 2D-3D correspondences in random trees based on the CRBNet descriptor. 
4) Experiments on challenging benchmarks show that the proposed localization method can significantly improve the robustness and accuracy compared with the ones which get nearest neighbor candidates of a given query local feature just by either local or global descriptors.

\section{Related Work}

\subsection{Visual Localization}
Indirect methods approximate the 6DoF camera pose of a given query image by using the pose of the image retrieved from the database. DenseVLAD \cite{torii201524} combines a densely sampled but compact global descriptor with the synthesis of new virtual views. NetVLAD \cite{arandjelovic2016netvlad} uses learned global descriptors while FAB-MAP \cite{cummins2008fab} uses Bag-of-Words paradigm to help image retrieval. Direct methods usually establish correspondences between 2D features in a given query image and 3D points in scene models, typically reconstructed using structure-from-motion (SfM) \cite{schoenberger2016sfm}. These 2D-3D correspondences are then used to estimate the pose of the query image. In order for practical use, 
some direct methods \cite{liu2017efficient} use co-visibility information to distinguish which parts of the scene are more likely visible in a given query image. Sv{\"a}rm et al. \cite{svarm2016city} first relax the matching criteria and then use deterministic outlier filter to handle numerous outliers resulting from the previous relaxed matching. This tactic is powerful especially when the gravity direction is known.
Sattler et al. \cite{sattler2016efficient} employ prioritized search strategy to improve efficiency.
Feng et al. \cite{feng2015fast} exploit label information in local descriptors of 3D points to train random trees which then is used to index 3D points in databases.

More and more serial indirect-direct methods use learning based global descriptors \cite{arandjelovic2016netvlad, passalis2017neural} to perform image retrieval recently and the scope of nearest neighbor candidates of a given query local feature is limited in the retrieval results. 
While direct methods only consider local descriptor information when getting the range of nearest neighbor candidates of a given query local feature.
We propose a novel framework of searching nearest neighbor candidates of a given query local feature by both local and global descriptors in parallel, which combines both of their merits.

\subsection{Learning Based Descriptor}
Although early hand-crafted local descriptors focused on gradient and intensity knowledge have been widely used, such as SIFT \cite{lowe2004distinctive}, ORB \cite{rublee2011orb}, BRISK \cite{leutenegger2011brisk}, many researchers have attempted to design the learned counterparts with the increasing popularity of machine learning and emergence of annotated patch benchmarks \cite{brown2010discriminative}.

It seems that end-to-end descriptors learned from patches by CNNs have become a tendency in recent years. MatchNet \cite{han2015matchnet} consists of a deep convolutional network that extracts descriptors from patches and a network of three fully connected layers that calculates a similarity between the extracted descriptors. DeepDesc \cite{simo2015discriminative} adopts Siamese network and an aggressive mining strategy towards hardly classified patches while DeepCompare \cite{Zagoruyko2015LearningTC} studies several neural network architectures to improve the performance. TFeat \cite{balntas2016learning} utilizes triplets of training samples and in-triplet mining of hard negatives. HardNet \cite{mishchuk2017working} minimizes distances of matching pairs, and meanwhile maximizes distances between patches and their “hardest-within-batch” negatives. GeoDesc \cite{luo2018geodesc} integrates geometry constraints from multi-view reconstructions to the learning process and provides guidelines of using learned local descriptors in structure-from-motion pipelines. SOSNet \cite{sosnet2019cvpr} incorporates a second order similarity regularization into its training process and designs a local descriptor evaluation method based on von Mises-Fischer distribution. L2-Net \cite{Tian_2017_CVPR} takes descriptor compactness into account. DOAP \cite{He_2018_CVPR} achieves better results by directly optimizing an average precision metric.

In addition, SuperPoint \cite{detone2018superpoint}, D2-Net \cite{dusmanu2019d2}, and R2D2 \cite{r2d2} are the representatives of joint detection and description methods. They just need a whole image as input, then keypoints and their corresponding descriptors are extracted by a relatively complex network. 

Although most learning based local descriptors have achieved good performances in certain tasks, they are real-valued descriptors and cannot be embedded in random trees directly.
Therefore, we propose a weighted Hamming regularization term to keep discriminativeness of learning based local descriptors after binarization.



\section{Visual Localization with CRBNet, Random Trees, and Parallel Search Framework}

CRBNet is to extract local descriptors for image points. We at first introduce CRBNet and then the parallel search framework.

\subsection{CRBNet}
\subsubsection{Network Architecture}
\begin{figure}
	\centering
	\centerline{\includegraphics[width=5.0cm]{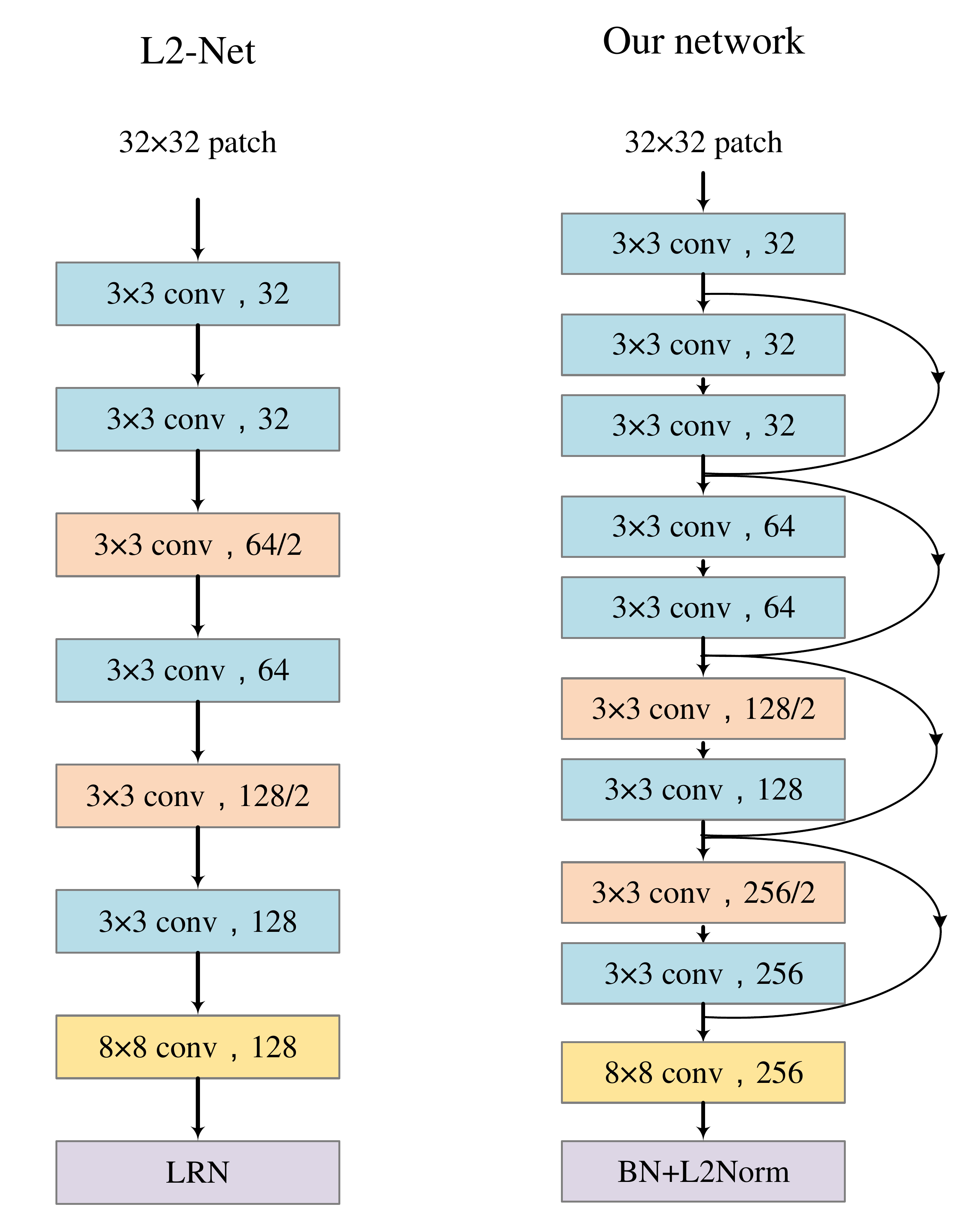}}
	\caption{Architectures of L2-Net (left) and our network (right). (Best viewed in color)}
	\label{network}
\end{figure}

Our network is based on L2-Net \cite{Tian_2017_CVPR}, which performs downsampling directly by strided convolutional layers and employs no pooling layer. 
Inspired by \cite{He_2016_CVPR}, we improve this network by adding residual learning framework. The overview of our network architecture is depicted in Fig. \ref{network}. The differences of our network from L2-Net are as follows. Except for the first and last layers, there are four stages by adding shortcut connections in our network.
Moreover, we put the convolutional layer with the stride of 2 into the last two stages in order to preserve the information of input patches with the size of 32\( \times \)32 as much as possible. Also, we use combination of batch normalization and L2-norm. The output of our network is a 256-D vector with unit-length.  

\subsubsection{Loss Function}
HardNet \cite{mishchuk2017working} achieves good results by using triplet margin loss. SOSNet \cite{sosnet2019cvpr} employs a second order similarity regularization to further supervise the process of descriptor learning. Considering both of their advantages, we adopt the triplet margin loss (\(\mathcal{H}_d\)) and second order similarity regularization (\(\mathcal{R}_{sos}\)) as part of our loss function to supervise the learning of real-valued descriptor. For binarizing descriptors, a simple way is to take the element greater than zero as 1, otherwise 0.
But this will cause a large quantization loss. Therefore, we establish a new constraint as the third part in our loss function to minimize the quantization loss.

Let \( (\textbf{\textit{x}}_i, \textbf{\textit{x}}_i^{+}) \) denote a pair of corresponding descriptors and let \( \mathcal{X} = (\textbf{\textit{x}}_i, \textbf{\textit{x}}_i^{+}), \\i=1...N \) denote the N pairs. The triplet margin loss is as:
\begin{equation}\label{1}
	\begin{split}
		\mathcal{H}_d = \frac{1}{N}\sum_{i=1}^{N}\max(0,1+d(\textbf{\textit{x}}_i, \textbf{\textit{x}}_i^{+})- \min_{\forall j, j\neq i}(d(\textbf{\textit{x}}_i, \textbf{\textit{x}}_j^{+}),d(\textbf{\textit{x}}_i^{+}, \textbf{\textit{x}}_j)))
	\end{split}
\end{equation}
where \( d \) is \( L_2 \) distance which measures the similarity of real-valued descriptors. 

The second term in our loss function is the second order similarity regularization as:
\begin{equation}\label{sos}
	\mathcal{R}_{sos} = \dfrac{1}{N}\sum_{i=1}^{N}\sqrt{\sum_{i \neq j}^{N} ((d(\textbf{\textit{x}}_i, \textbf{\textit{x}}_j) - d(\textbf{\textit{x}}_i^{+}, \textbf{\textit{x}}_j^{+}) )^2)}
\end{equation}
where \( N \) is the number of pairs in a batch, \( i,j \in [1,...,N] \).

Our new regularization term is still in the form (\ref{1}) but the metric $d$ is replaced by a weighted Hamming distance \( d_w \) :
\begin{equation}\label{2}\!
	d_w(\textbf{\textit{x}}_i, \textbf{\textit{x}}_i^{+}) \! =\!\sqrt{\sum_{k=1}^{K} {((\textbf{\textit{x}}_{ik} - \textbf{\textit{x}}_{ik}^{+})^{2} \times (\rm{sign}(\textbf{\textit{x}}_{\it{ik}}) \!-\! \rm{sign}(\textbf{\textit{x}}_{\it{ik}}^{+}) ) )}}
\end{equation}
where \( K \) is the descriptor dimension, $k \in [1,...,K] $, $i$ denotes pair index, and \( \rm{sign}(\textbf{\textit{x}}_\textit{i}) \) is the binary descriptor. So \( (\textbf{\textit{x}}_{ik} - \textbf{\textit{x}}_{ik}^{+})^{2} \) can be regarded as a weight of the $k$-th dimension when calculating Hamming distance. This term is used to supervise the learning of binary descriptors. 

When it comes to learn binary descriptors, the first thing that comes to mind is Hamming distance. One possible way to learn a binary descriptor is to implement a sign function to binarize the descriptor and then use a standard Hamming distance. Unfortunately, the sign activations cannot propagate the gradient backwards. Some previous methods \cite{cao2017hashnet} try to change the sign function to solve this problem which introduce some extra hyperparameters. We propose a weighted Hamming regularization where the weight is associated with the real-valued descriptor and differentiable. An intuitive explanation is: when the $k$-th elements of the matching descriptors have the same sign, this regularization will not give penalty. Otherwise, the regularization penalizes according to the difference between the corresponding real-valued descriptors. The most similar method to our proposed regularization is BinGAN \cite{zieba2018bingan}. The difference is that BinGAN imposes weights on the dot product of different binary vectors in a batch to increase diversity of binary vectors, while we utilize the weight on dimension-wise of Hamming distance to learn similarity between matching patches. 

Our total loss function is the sum of the above three terms:
\begin{equation}\label{3}
	\mathcal{L} = \mathcal{H}_d + \mathcal{R}_{sos} + \mathcal{H}_{d_w}
\end{equation}


\subsection{Parallel Search Framework}
\subsubsection{Overview} 
\begin{figure*}
	\centering
	\centerline{\includegraphics[width=\textwidth]{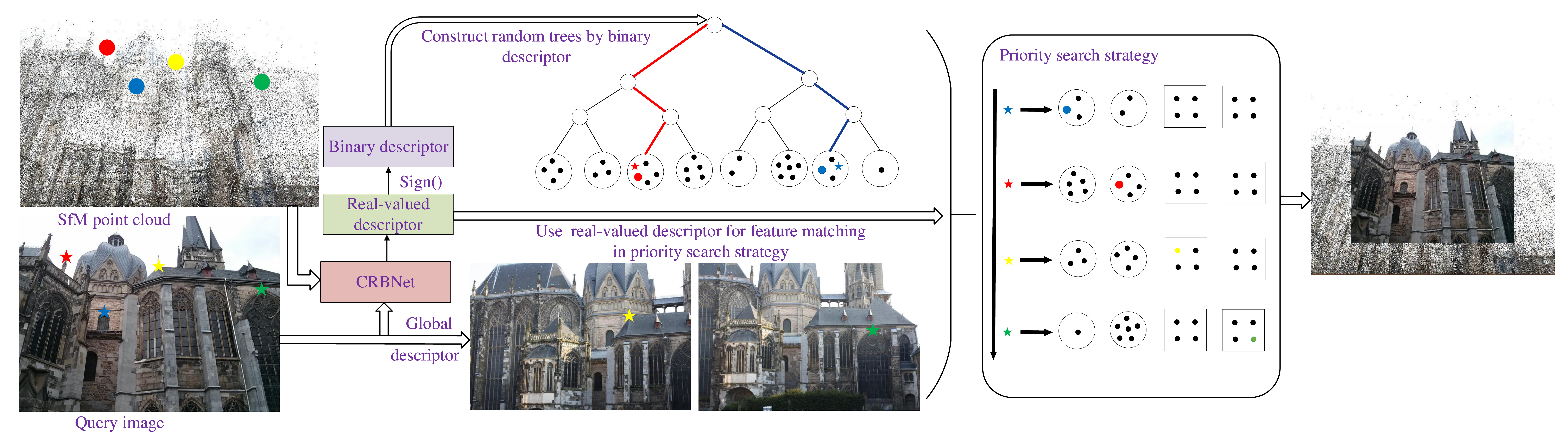}}
	\centering
	\caption{An overview of the proposed framework for visual localization. $\star$ means a local feature in a given query image and $ \bullet $ means 3D points in databases. $ \bullet $ is the nearest neighbor of $\star$ if they are the same color. In priority search strategy, $ \bigcirc $ indicates prior leaf nodes obtained by a probability model in the random trees constructed by supervised information of local descriptors and $ \square $ indicates prior frames obtained by global descriptors. Then correspondences found in prior leaf nodes and prior frames are used to compute the 6DoF pose of the given query image. (Best viewed in color)}
	\label{framework}
\end{figure*}

An overview of the proposed framework is shown in Fig. \ref{framework}. At first, a 3D point cloud is constructed by SfM \cite{schoenberger2016sfm}. Local descriptors for these 3D points on their corresponding database images and for query points in a given query image are extracted by the proposed CRBNet. 
The binary CRBNet is used to construct random tree structures for clustering 3D points. Meanwhile, 3D points visible by a database image are also clustered into one frame. For different images, we obtain different frames.
Prior leaf nodes in random trees are obtained by a probability model while prior frames are obtained by a global search.
These indicate the nearest neighbor candidates of a given query local feature in prior leaf nodes obtained by local information and in prior frames by global information. Then, the real-valued CRBNet is used to compute Euclidean distances between the query local feature and its nearest neighbor candidates. The 3D point with the minimum distance is the correspondence of the 2D query point while the distance ratio test \cite{lowe2004distinctive} is passed. By the same way, all 2D query points in the given image are matched to their corresponding 3D points. Finally, the correspondences between these 2D query local features and 3D points are used to compute the 6DOF camera pose of the given image by PnP and RANSAC algorithms. 

\subsubsection{Motivation} Visual localization remains a challenging task in large-scale environments. In outdoor scenes, images of the same place can have significant differences because of changes on illumination, season, viewpoint, blur, etc. Also, some places like buildings have a lot of repetitive textures which undoubtedly increase difficulty of visual localization. Existing localization framework cannot perform well in all these situations. 

These days, CNNs are increasingly adopted in serial indirect-direct methods to extract global descriptors of database images and a given query image. Then Euclidean distances are calculated between global descriptors of database images and the given query image. The smaller the distance is, the more similar the database image to the query image is. Serial indirect-direct methods just search nearest neighbors of a given query local feature among the 3D points visible by KNNs in databases. Therefore, the localization results of serial indirect-direct methods seriously depend on the retrieved images. When the viewpoint of the given query image changes largely, the global descriptor of the given query image and its KNNs are likely different, which would cause incorrect retrieved images.
Feng et al. \cite{feng2015fast}, 
only depending on supervised information among local descriptors, cluster 3D points to leaf nodes of random trees. The 3D points in one leaf node can be visible to different database images. So generally speaking, this method is more robust to large viewpoint changes. However, when illumination is greatly various, like images taken in daytime and nighttime, descriptors extracted from matching local patches will have large difference. Under this circumstance, only depending on descriptors extracted from local patches to cluster 3D points is unreliable and using global descriptors is a wise choice. 
Parallel strategy which utilizes the local and global descriptors simultaneously to find nearest neighbor candidates of a given query local feature has a promise to perform well in various conditions. 

\begin{figure*}
	\centering
	\subfigure {
		\begin{minipage}[t]{0.23\linewidth}
			\centering
			\includegraphics[width=\linewidth]{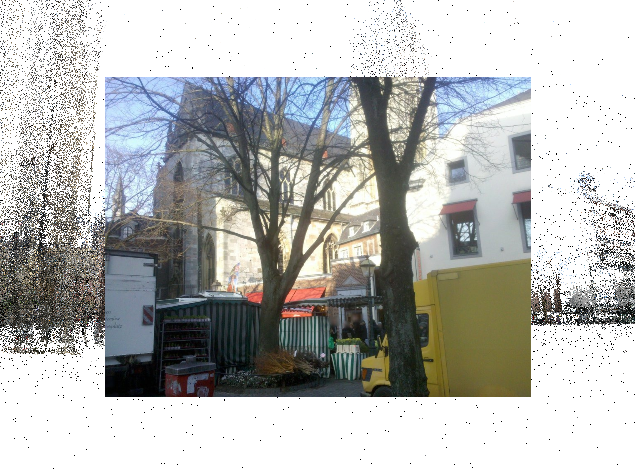}
	\end{minipage}}
	\subfigure {
		\begin{minipage}[t]{0.23\linewidth}
			\centering
			\includegraphics[width=\linewidth]{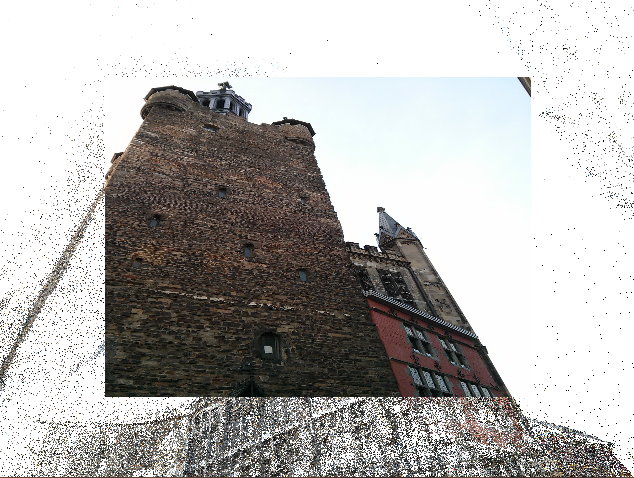}
	\end{minipage}}
	\subfigure {
		\begin{minipage}[t]{0.23\linewidth}
			\centering
			\includegraphics[width=\linewidth]{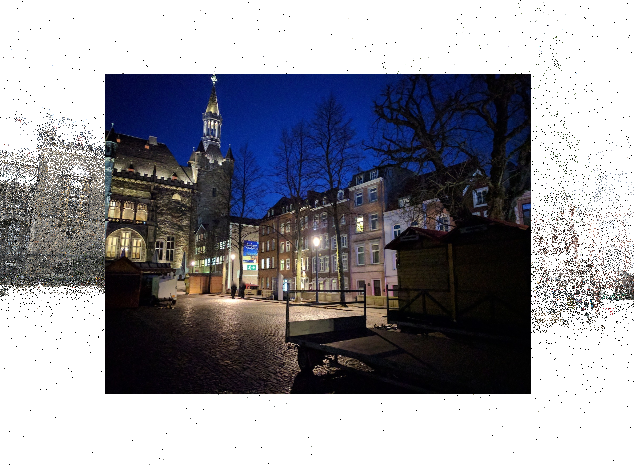}
	\end{minipage}}
	\subfigure {
		\begin{minipage}[t]{0.23\linewidth}
			\centering
			\includegraphics[width=\linewidth]{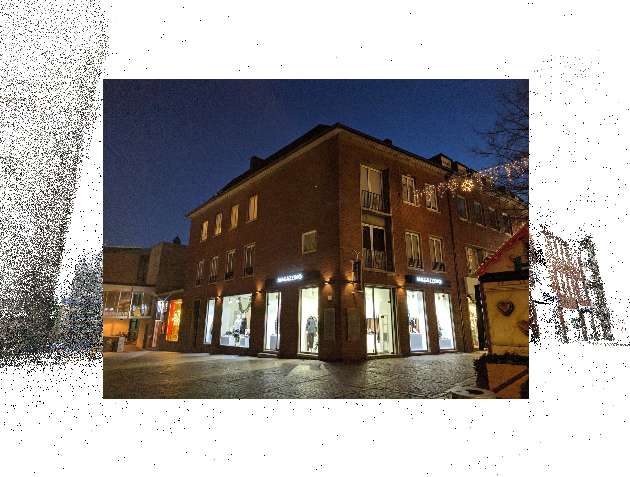}
	\end{minipage}}
	
	\subfigure {
		\begin{minipage}[t]{0.23\linewidth}
			\centering
			\includegraphics[width=\linewidth]{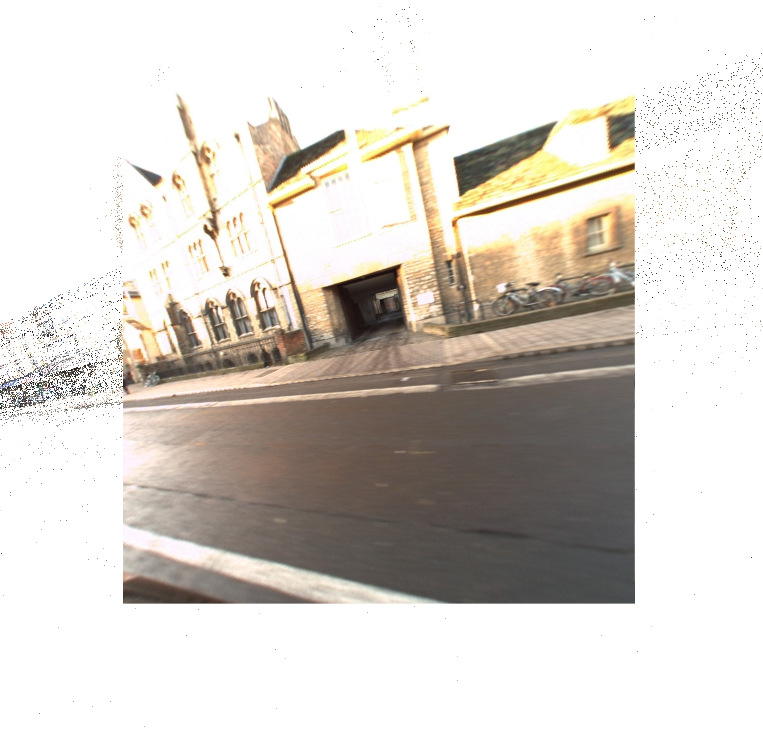}
	\end{minipage}}
	\subfigure {
		\begin{minipage}[t]{0.23\linewidth}
			\centering
			\includegraphics[width=\linewidth]{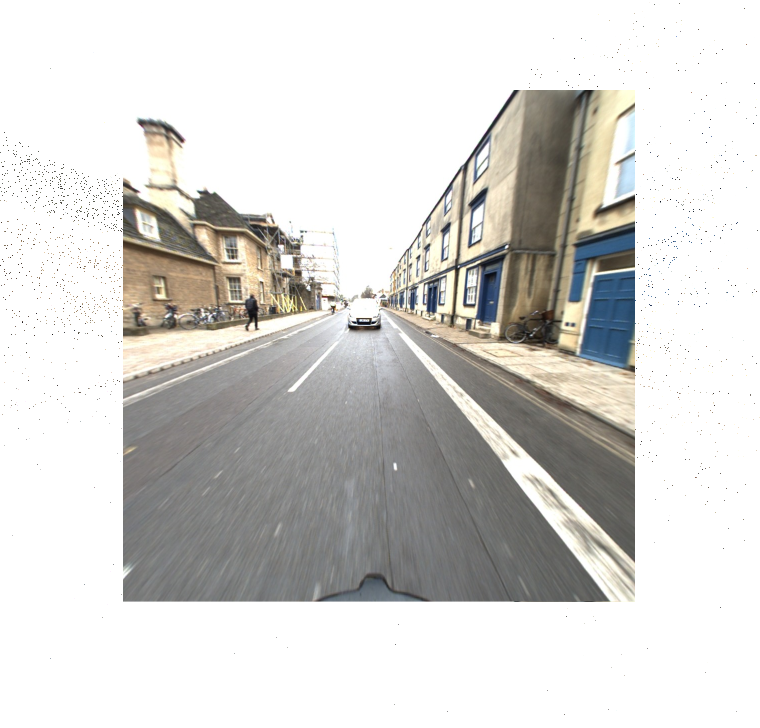}
	\end{minipage}}
	\subfigure {
		\begin{minipage}[t]{0.23\linewidth}
			\centering
			\includegraphics[width=\linewidth]{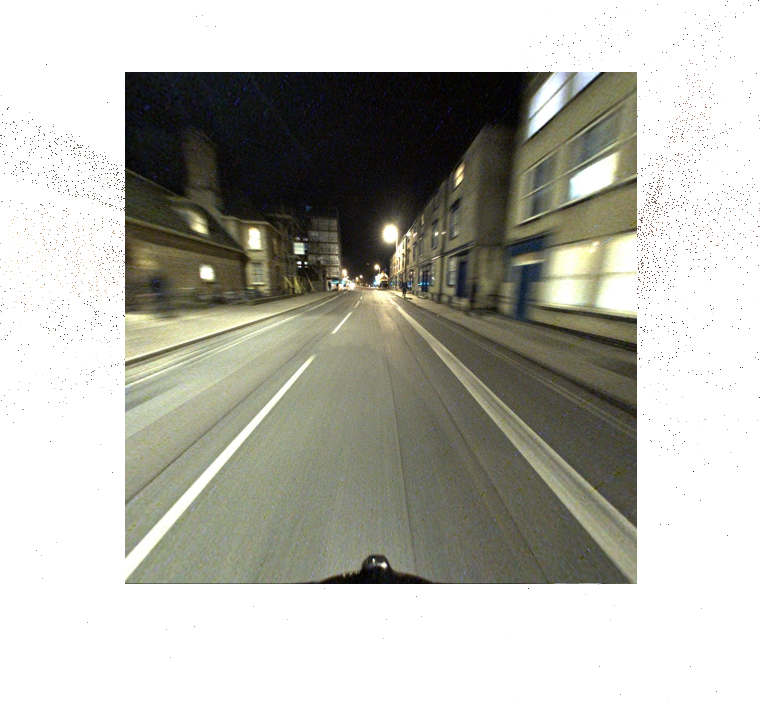}
	\end{minipage}}
	\subfigure {
		\begin{minipage}[t]{0.23\linewidth}
			\centering
			\includegraphics[width=\linewidth]{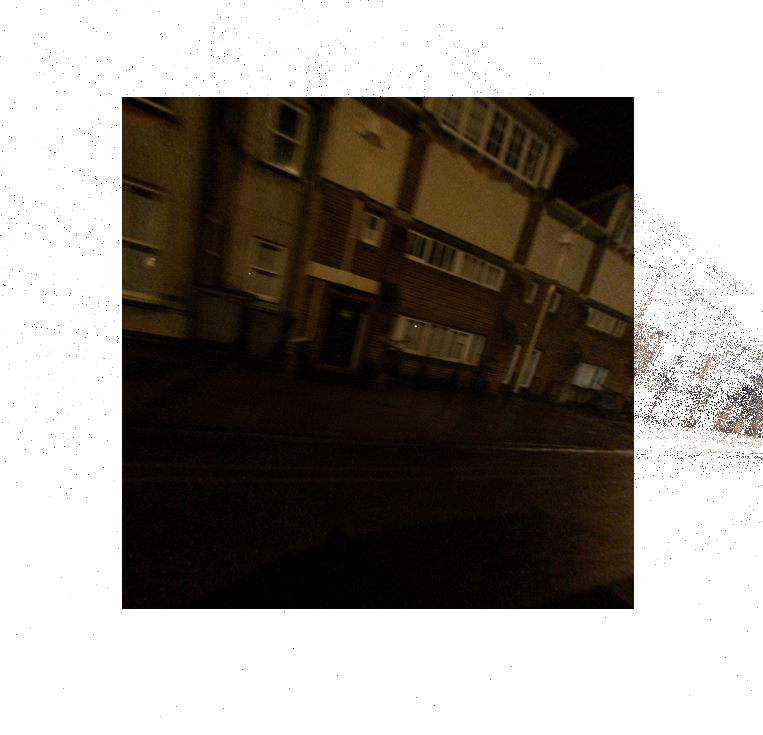}
	\end{minipage}}
	
	\caption{Some examples successfully localized by the proposed framework, where images of the first line are from Aachen Day-Night dataset and images of the second line are from RobotCar Seasons dataset \cite{sattler2018benchmarking}. (Best viewed in color)}
	\label{parallel_search_example}
\end{figure*}

Based on the above analysis, we propose a new framework which takes advantages of both local and global descriptors when finding nearest neighbor candidates of a given query local feature. 
We use random tree structures constructed by local descriptors to cluster 3D points in parallel search framework. 
When constructing the random trees, local descriptors are binarized. We also propose a weighted Hamming regularization term to keep discriminativeness of local descriptors after binarization, as illustrated in Section 3.1.
This framework can effectively localize the query image in various situations. Some examples localized by the proposed framework are shown in Fig. \ref{parallel_search_example}.

\subsubsection{Prior Leaf Nodes}
Random trees have been used to index binary descriptors \cite{feng2015fast}. The leaf nodes of random trees are used for storing the indices of 3D points in databases while each non-leaf node selects one dimension of binary descriptors determining which leaf node the 3D points are stored in. Once the value in the non-leaf node is fixed, a descriptor in the database traverses the tree starting from the root node, then turning to either the right or left child node based on whether the value in current node is one or zero until arriving at a leaf node.
After that the descriptor and its corresponding 3D point are placed in the arrived leaf node. When searching the approximate nearest neighbor of a given query local feature, the query local feature needs to find the leaf node that it belongs to in the same way of indexing these 3D points. Then we conduct a linear search in the found leaf node in which the query local feature's nearest neighbor is most likely stored. In order to improve the performance of approximate nearest neighbor search, multiple trees are adopted in this paper. 
We use the binary form of CRBNet to construct random trees while the real-valued descriptor for the linear search in leaf nodes. 
The speed of approximate nearest neighbor search based on random trees is very fast compared with hierarchical clustering trees (HCT), because traversing random trees does not need any distance calculation.

To further improve the accuracy, we conduct the linear search in multiple leaf nodes of one random tree. Feng et al. \cite{feng2015fast} proposes a priority search strategy, where the probability model is very crucial because the model can distinguish which leaf node most likely includes the nearest neighbor of a given query local feature. However the probability model in \cite{feng2015fast} is suitable for traditional binary descriptors. In the following, we present a probability model which is suitable for the proposed learning based local descriptors. 
Due to the variance of illumination and viewpoint when intercepting patches, the nearest neighbor  \( \rm\textbf{p} \) of a given query real-valued descriptor \( \rm\textbf{q} \) can be treated as the result of applying a perturbation \( \Delta \) to \( \rm\textbf{q} \), \( \textit{i.e. }  \rm\textbf{p} = \rm\textbf{q} + \Delta \). So the probability that \( \rm\textbf{p} \) and \( \rm\textbf{q} \) are included in the same leaf node \( \textbf{\textit{l}} \) can be denoted as (\ref{4}):
\begin{equation}\label{4}
	\begin{split}
		\textit{P}(\tau(\rm\textbf{p}) = \textbf{\textit{l}} | \rm\textbf{p} = \rm\textbf{q} + \Delta ) =  \prod_{\it{k} = \rm{1}}^K \textit{P}(\tau_{\it{k}}(\textit{p}_{\it{d}})=\textit{l}_{\it{k}} | {\textit{p}}_{\it{d}} = {\textit{q}}_\textit{d} + \delta_\textit{d})
	\end{split}
\end{equation}
where \( \rm{\it{K}} \) is the depth of random trees, $p_d$ is the $d$-th element of ${\bf p}$, and \( \tau_{\rm{\it{k}}}(p_d) \) is the $d$-th element of the binary descriptor \( \textbf{p} \). Please notice that $d$ may be different in different non-leaf node test, which depends on the supervised training process. $l_k \in (0, 1) $ is the $k$-th node test on the path to the leaf node \( \textbf{\textit{l}} \).  
$P(\tau_k(p_d)=l_k | p_d = q_d + \delta_d )$  
means the probability that the $d$-th element of the binary form of descriptor \( \textbf{p} \) equals to \( l_k \).

The key calculation for \( P(\tau_k(p_d)=l_k | p_d = q_d + \delta_d) \) is to calculate the distribution of \( \delta_d \). In fact, the perturbation between matching patches is a random variable with unknown distribution. The number of descriptors is usually extensive in 3D models, according to Central Limit Theorem, normal distribution can give the best approximation. We use normal distribution to describe the perturbation, \( \textit{i.e.}\) \( \delta_{\it{d}}  \sim \rm{N}( \mu_{\it{d}}, {\sigma_{\it{d}}}^{2} ) \). Then we know that \( p_d  \sim \rm{N}( \mu_{\it{d}} + \textit{q}_\textit{d}, {\sigma_{\it{d}}}^{2} ) \) and can calculate \( P(\tau_{\it{k}}(p_d)=l_k | p_d=q_d + \delta_d) \) as:
\begin{equation}\label{5}
	\begin{split}
		\left\{
		\begin{array}{lr}
			P(\tau_k(p_d)= 0 | p_d = q_d+\delta_d ) \! = \! \textit{P}(p_d < 0 |  p_d \! = \! q_d+\delta_d  )
			\\ \!=\! \int_{- \infty}^{0}\frac{1}{\sigma_{\it{d}} \sqrt{2\pi}}e^{-\frac{(p_d-(\mu_{\it{d}} + q_d ))^{2}}{2\sigma_{\it{d}}^2}}\rm{d} \textit{p}_{\it{d}} \\
			P(\tau_k(p_d) \! = \! 1 | p_d \! = \! q_d + \delta_d) \! = \! \textit{P}(p_d \! >= \! 0 | p_d  \!= \! q_d + \delta_d )
			\\= \int_{0}^{\infty}\frac{1}{\sigma_{\it{d}} \sqrt{2\pi}}e^{-\frac{(p_d-(\mu_d + q_d ))^{2}}{2\sigma_{\it{d}}^2}}\rm{d}\rm \textit{p}_\textit{d} 
		\end{array} 
		\right.
	\end{split}
\end{equation}
Now, we can get the probability \( P(\tau(\rm\textbf{p}) = \textbf{\textit{l}} | \rm\textbf{p} = \rm\textbf{q} + \Delta ) \) by bringing (\ref{5}) into (\ref{4}). In practice, \( \mu_{\it{d}} \) and \( \sigma_{\it{d}} \) may be different along with different node tests. For efficient and practical use, we adopt an uniform values \( \mu \) and \( \sigma \) which are the average of \( \mu_{\it{d}} \) and \( \sigma_{\it{d}} \) of 10 randomly selected tests, with each \( \mu_{\it{d}} \) and \( \sigma_{\it{d}} \) estimated from 100,000 samples.

By the above probability, we search the leaf node with the maximum probability, i.e. prior leaf node, for establishing 2D-3D correspondences in priority.

\subsubsection{Prior Frames}

Although prior frames can be obtained by bag-of-words \cite{karakasis2015image}, VLAD, Fisher vector or hybrid features \cite{li2018multi}, we do not adopt these traditional methods in that they are aggregated by local descriptors. Instead, we use NetVLAD \cite{arandjelovic2016netvlad} to extract global descriptors for database and a given query image, then a global search in database images is performed to find KNNs of the given query image. The KNNs are the prior frames that we want. The 3D points that are visible in the prior frames are the nearest neighbor candidates of a given query local feature.

After we find prior leaf nodes and prior frames, we put the 3D points in them together. These 3D points make up the whole nearest neighbor candidates of the given query local feature. We perform a linear search among the nearest neighbor candidates of a given query local feature by the real-valued form of CRBNet. In the process of getting nearest neighbor candidates of a given query local feature, parallel search framework fuses the information of both local and global descriptors, which can make the nearest neighbor candidates of a given query local feature comprehensive laying a solid foundation for computing accurate 6DoF camera pose of the query image.

Benefit from the parallel search framework, the localization method can find accurate 2D-3D correspondences even the query image is taken in extreme conditions largely different from database images, like nighttime, season changes, viewpoint changes and so on. Once the process of establishing the correspondences is finished, we use RansacLib \cite{Sattler2019Github,Lebeda2012BMVC} to solve poses of the query image.

\section{Experiments}

In this section, we will evaluate CRBNet on UBC Phototour dataset \cite{brown2010discriminative} and HPatches \cite{Balntas17} first in Section 4.1. UBC Phototour dataset and HPatches are two popular datasets for local descriptor evaluation. Then on the long term visual localization RobotCar Seasons dataset and Aachen Day-Night dataset \cite{sattler2018benchmarking}, the direct random tree localization method with our CRBNet and new probability model will be evaluated in Section 4.2, and in Section 4.3 the proposed parallel search framework will be evaluated. In order to make a fair comparison when evaluating all the localization methods, we don't train CRBNet in RobotCar Seasons dataset and Aachen Day-Night dataset, instead we train CRBNet only on the three subsets of UBC Phototour dataset. 


\subsection{Evaluation of CRBNet}
We use PyTorch library to train 100 epochs. Input of CRBNet is a patch with size of 32 \( \times \) 32 pixels. The batch size is 1024 and the optimization is done by Adam optimizer \cite{kingma2014adam} with \( \alpha \) = 0.01, \( \beta_1 \) = 0.9 and \( \beta_2 \) = 0.999. Similar to the previous work \cite{sosnet2019cvpr}, a dropout layer with 0.1 dropout rate is employed before the last convolutional layer and 8 nearest neighboring pairs are selected for a given pair to calculate \( \mathcal{R}_{sos} \).
\begin{table*}
	\caption{Patch verification performance on UBC Phototour dataset, where the metric is false positive rate at 95\% recall. The second column shows the dimensions of descriptors and suffix "+" indicates data augmentation. The best results are in bold. We can observe that our descriptor outperforms all other methods both in binary and real-valued form in most items.}
	\begin{center}
		\begin{tabular}{ p{2.78cm}<{\centering} | p{0.7cm}<{\centering} | p{0.6cm}<{\centering}  p{0.6cm}<{\centering} |p{0.6cm}<{\centering} p{0.6cm}<{\centering}| p{0.6cm}<{\centering} p{0.6cm}<{\centering} | p{1.3cm}<{\centering}} 
			\hline
			\multirow{2}{*}{Method} & Train & Not & Yos & Lib & Yos & Lib & Not & \multirow{2}{2.5em}{mean FPR95}\\
			\cline{2-8}
			&Test& \multicolumn{2}{c|}{Lib} & \multicolumn{2}{c|}{Not} & \multicolumn{2}{c|}{Yos} & \multicolumn{1}{c}{} \\
			\hline
			\multicolumn{9}{c}{\textit{Real-valued descriptors}}\\
			\hline
			SIFT \cite{lowe2004distinctive}& 128 & \multicolumn{2}{c|}{29.84} & \multicolumn{2}{c|}{22.53} & \multicolumn{2}{c|}{27.29} & 26.55 \\
			TFeat+ \cite{balntas2016learning} & 128 & 7.39 & 10.13 & 3.06 & 3.80 & 8.08 & 7.24 & 6.64 \\
			L2Net+ \cite{Tian_2017_CVPR} & 128 & 2.36 & 4.70 & 0.72 & 1.29 & 2.57 & 1.71 & 2.23 \\
			CS L2Net+ \cite{Tian_2017_CVPR} & 256 & 1.71 & 3.87 & 0.56 & 1.09 & 2.07 & 1.30 & 1.76 \\
			HardNet+  \cite{mishchuk2017working} & 128 & 1.49 & 2.51 & 0.53 & 0.78 & 1.96 & 1.84 & 1.51 \\
			HardNet-GOR+  \cite{mishchuk2017working} & 128 & 1.48 & 2.43 & 0.51 & 0.78 & 1.76 & 1.53 & 1.41 \\
			DOAP+ \cite{He_2018_CVPR} & 128 & 1.54 & 2.62 & 0.43 & 0.87 & 2.00 & 1.21 & 1.45 \\
			SOSNet+  \cite{sosnet2019cvpr} & 128 & 1.08 & 2.12 & 0.35 & 0.67 & \textbf{1.03} & 0.95 & 1.03 \\
			\textbf{CRBNet+}  & 256 & \textbf{0.99} & \textbf{1.79} & \textbf{0.33} & \textbf{0.57} & 1.24 & \textbf{0.93} &\textbf{0.98} \\
			\hline
			\multicolumn{9}{c}{\textit{Binary descriptors}}\\
			\hline
			BinBoost+ \cite{trzcinski2013boosting} & 64 & 20.49 & 21.67 & 16.90 & 14.54 & 22.88 & 18.97 & 19.24 \\
			L2Net \cite{Tian_2017_CVPR} & 128 & 10.3 & 11.71 & 6.37 & 6.76 & 13.5 & 11.57 & 10.03 \\
			SOSNet+ITQ  & 128 & 8.49 & 11.4 & 4.87 & 6.06 & 10.35 & 8.43 & 8.26 \\
			L2Net+ \cite{Tian_2017_CVPR} & 128 & 7.44 & 10.29 & 3.81 & 4.31 & 8.81 & 7.45 & 7.01 \\
			CRBNet+  & 128 & 5.92 & 8.73 & 3.06 & 3.80 & 8.23 & 6.71 & 6.08 \\
			CS L2Net+ \cite{Tian_2017_CVPR} & 256 & 4.01 & 6.65 & 1.90 & 2.51 & 5.61 & 4.04 & 4.12 \\
			DOAP+ \cite{He_2018_CVPR} & 256 & 3.18 & 4.32 & \textbf{1.04} & \textbf{1.57} & \textbf{4.10} & 3.87 & 3.01 \\
			
			CRBNet(N)+  & 256 & 4.13 & 5.99 & 3.11 & 2.18 & 7.78 & 4.07 & 4.54 \\
			\textbf{CRBNet+}  & 256 & \textbf{2.83} & \textbf{4.28} & 1.38 & 1.60 & 4.33 & \textbf{3.45} & \textbf{2.98} \\
			\hline
		\end{tabular}
	\end{center}
	\label{table_desc}
\end{table*}

UBC Phototour dataset \cite{brown2010discriminative} is one of the most popular datasets for local descriptor learning. It contains three subsets: \textit{Liberty}, \textit{Notre Dame}, and \textit{Yosemite}.
We follow the standard evaluation protocol of \cite{brown2010discriminative} and train our models on one subset and test on the other two. Then we report the false positive rate at 95\% recall in the 100K pairs provided by \cite{brown2010discriminative}. For the real-valued descriptors, we compare our descriptor with the state-of-art descriptors, including TFeat \cite{balntas2016learning}, L2Net \cite{Tian_2017_CVPR}, HardNet \cite{mishchuk2017working}, DOAP \cite{He_2018_CVPR}, SOSNet \cite{sosnet2019cvpr}, on UBC Phototour dataset \cite{brown2010discriminative}. For the binary part, we compare binary CRBNet with BinBoost \cite{trzcinski2013boosting}, binary L2Net \cite{Tian_2017_CVPR}, DOAP \cite{He_2018_CVPR}, SOSNet \cite{sosnet2019cvpr}. As SOSNet is a real-valued descriptor, we use ITQ \cite{gong2012iterative} to create binary parts. ITQ uses an iterative approach to minimize the binarization error when mapping real-valued descriptors to binary descriptors and we refer the readers to \cite{gong2012iterative} for the details.  We use the network of L2Net to produce 128-D outputs while 256-D outputs are based on the network described on Section 3.1.  CRBNet(N) means the weighted Hamming distance regularization is not used.

Comparison results are shown in Table \ref{table_desc}, where the results of SIFT provided by \cite{brown2010discriminative} are regarded as a baseline. For the binary part, we can see the binary descriptors of 256-D are better than the ones of 128-D. That is to say that increasing output dimension can improve performance of the binary descriptor. No matter the output is 128-D or 256-D, our binary descriptors get the best results compared with the same dimension methods. We also implement an ablation study, i.e., we remove the weighted Hamming distance from CRBNet, and then get CRBNet(N), its results are also listed in Table \ref{table_desc}. Compared with CRBNet(N), the performance of CRBNet gets a significant boost with the help of weighted Hamming network. Last but not least, our method achieves the best performance in binary form meanwhile still keeps excellent performance in real-valued form for most cases.

\begin{figure*}
	\vspace{0.5cm}
	\centering
	\centerline{\includegraphics[width=\textwidth]{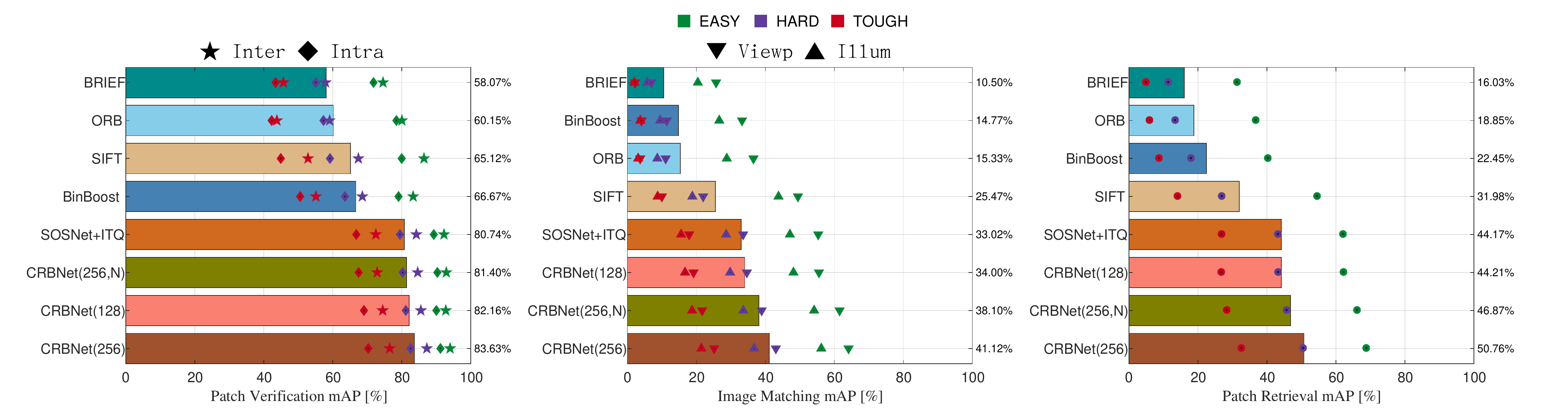}}
	\centering
	\caption{Verification, matching and retrieval results on HPatches \cite{Balntas17}. All the descriptors are tested on split 'a' and trained on \textit{Liberty} subset of UBC Phototour dataset \cite{brown2010discriminative} without augment. (Best viewed in color)}
	\label{HPatches}
\end{figure*}

In order to further evaluate our weighted Hamming loss and network, we also evaluate the binary CRBNet on HPatches dataset \cite{Balntas17}. 
This dataset contains 116 sequences and every sequence consists of 6 images. 69 of these sequences have great variations in viewpoint while the others have great changes in illumination. Keypoints in HPatches are detected by DoG, Harris and Hessian detectors. In each of the sequences, according to level of geometric noise, matching patches are divided into three groups: \textit{Easy, Hard,} and \textit{Tough}. HPatches dataset also defines three evaluation tasks: patch verification, image matching, and patch retrieval. In this benchmark, we compare the binary part with ORB\cite{rublee2011orb}, BinBoost \cite{trzcinski2013boosting},  SIFT \cite{lowe2004distinctive} and SOSNet+ITQ. Because we don't find the pretrained model of DOAP and CS L2Net, so we cannot compare binary CRBNet with them in HPatches. The results of three tasks are shown in Fig. \ref{HPatches}. As we can see, for 128-D, our CRBNet is better than SOSNet+ITQ; For 256-D, CRBNet with weighted Hamming distance regularization is better than the one without weighted Hamming distance regularization, which indicates that the weighted Hamming distance regularization is effective when learning binary descriptors.

\subsection{Evaluation of the Direct Random Tree Method}

At first, let's look at behavior of the direct random tree method \cite{feng2015fast} while with our CRBNet and new probability model. This method is denoted as RT\_AP+CRBNet. Please notice that RT\_AP+CRBNet is a pure direct method at the moment and just uses the supervised information of local descriptors to select nearest neighbor candidates of a given query local feature.
We use 6 random trees with depth of 23 and check top 100 leaf nodes in the priority search strategy.

The long-term visual localization RobotCar Seasons dataset and Aachen Day-Night dataset \cite{sattler2018benchmarking} are used in this section. The former dataset, including 20,862 reference images, 6.77M 3D points and 11,934 query images, is based on a subset of the Oxford RobotCar dataset \cite{RobotCarDatasetIJRR}. 
Some images are also badly blurry because of motion which adds extra difficulties for accurate visual localization. 
The latter dataset, containing 4,328 reference images and 1.65M 3D points, is based on the Aachen localization dataset from \cite{sattler2012image}.
In query image sets, 824 images are collected during daytime while the other 98 images are collected during nighttime.
The query images are captured by different types of cameras which makes it more difficult to localize. 
We report the pose recall at different position and orientation thresholds for query images. The thresholds are the same as \cite{sattler2018benchmarking}.

\begin{table*}[h]
	\caption{Localization evaluation on RobotCar Seasons dataset. The methods in the last two lines are ours. RT\_AP+CRBNet is a direct method with our CRBNet. ParallelSearch+CRBNet in the last line is ours of the new parallel search framework. The best results are in bold and the second-best
		are in red.}
	\begin{center}
		\resizebox{\textwidth}{15mm}
		{
			\begin{tabular}{ c |c |c |c |c |c |c |c || c | c } 
				
				& \multicolumn{7}{c||}{\textbf{day conditions}} & \multicolumn{2}{c}{\textbf{night conditions}} \\
				&dawn&dusk&OC-summer&OC-winter&rain&snow&sun&night&night-rain\\
				\cline{2-10}
				\multirow{2}{*}{\shortstack{m\\deg} }
				& \multirow{2}{*}{\shortstack{.25/.50/5.0\\2/5/10} } &\multirow{2}{*}{\shortstack{.25/.50/5.0\\2/5/10} } &\multirow{2}{*}{\shortstack{.25/.50/5.0\\2/5/10} } &\multirow{2}{*}{\shortstack{.25/.50/5.0\\2/5/10} } &\multirow{2}{*}{\shortstack{.25/.50/5.0\\2/5/10} } &\multirow{2}{*}{\shortstack{.25/.50/5.0\\2/5/10} } &\multirow{2}{*}{\shortstack{.25/.50/5.0\\2/5/10} }
				&\multirow{2}{*}{\shortstack{0.5/.50/5.0\\2/5/10} }
				&\multirow{2}{*}{\shortstack{0.5/.50/5.0\\2/5/10} }\\
				&\multicolumn{1}{c|}{ } & \multicolumn{1}{c|}{ }& \multicolumn{1}{c|}{ }& \multicolumn{1}{c|}{ }& \multicolumn{1}{c|}{ }& \multicolumn{1}{c|}{ }& \multicolumn{1}{c||}{ }& \multicolumn{1}{c|}{ }& \multicolumn{1}{c}{ }\\
				\hline
				Active Search\cite{sattler2016efficient}&45.3 / 76.0 / 93.0&52.0 / 83.0 / 95.9&32.8 / 74.1 / 97.8&37.4 / 78.7 / 94.6&58.4 / 82.4 / 97.6&50.5 / 81.8 / 95.5&29.6 / 57.4 / 84.1&1.6 / 3.9 / 10.5&2.0 / 10.9 / 18.0\\
				\hline
				CSL\cite{svarm2016city}&47.2 /73.3 / 90.1&56.6 / 82.7 / 95.9&34.1 / 71.1 / 93.5&39.5 / 75.9 / 92.3&59.6 / 83.1 / 97.6&53.2 / 83.6 / 92.4&28.0 / 47.0 / 70.4&0.2 / 0.9 / 5.3&0.9 / 4.3 / 9.1\\
				\hline
				DenseVLAD\cite{torii201524}&8.7 / 36.9 / 92.5&10.2 / 38.8 / 94.2&6.0 / 29.8 / 92.0&4.1 / 26.9 / 93.3&10.2 / 40.6 / 96.9&8.6 / 30.1 / 90.2&5.7 / 16.3 / 80.2&0.9 / 3.4 / 19.9&1.1 / 5.5 / 25.5\\
				\hline
				NetVLAD\cite{arandjelovic2016netvlad}&6.2 / 22.8 / 82.6&7.4 / 29.7 / 92.9&6.5 / 29.6 / 95.2&2.8 / 26.2 / 92.6&9.0 / 35.9 / 96.0&7.0 / 25.2 / 91.8&5.7 / 16.5 / 86.7&0.2 / 1.8 / 15.5&0.5 / 2.7 / 16.4\\
				\hline
				FABMAP\cite{cummins2008fab}&1.2 / 5.6 / 14.9&4.1 / 18.3 / 55.1&0.9 / 8.9 / 39.3&2.6 / 13.3 / 44.1&8.8 / 32.1 / 86.5&2.0 / 8.2 / 28.4&0.0 / 0.0 / 2.4&0.0 / 0.0 / 0.0&0.0 / 0.0 / 0.0\\
				\hline
				NV+SP \cite{sarlin2019coarse}&52.8 / 79.1 / 95.4&55.1 / 83.5 / 95.9&\textcolor{red}{\textbf{43.8}} / \textcolor{red}{\textbf{78.2}} / 98.7&\textcolor{red}{\textbf{50.8}} / 78.2 / 95.9&62.0 / \textbf{84.8} / 96.9&\textbf{60.5} / 85.1 / 95.7&\textbf{52.0} / \textcolor{red}{\textbf{74.3}} / \textcolor{red}{\textbf{93.3}}&\textbf{11.9} / \textbf{24.7} / \textbf{48.9}&\textcolor{red}{\textbf{10.5}} / \textbf{30}.7 / \textbf{49.3}\\
				\hline
				\textbf{RT\_AP+CRBNet}&\textcolor{red}{\textbf{53.8}} / \textcolor{red}{\textbf{79.5}} / \textcolor{red}{\textbf{96.3}}&\textcolor{red}{\textbf{59.4}} / \textbf{84.3} / \textbf{98.2}&39.5 / \textcolor{red}{\textbf{78.2}} / \textbf{99.8}&42.6 / \textcolor{red}{\textbf{81.0}} / \textcolor{red}{\textbf{98.2}}&\textbf{63.2} / \textcolor{red}{\textbf{83.8}} / \textbf{99.3}&57.1 / \textcolor{red}{\textbf{85.3}} / \textbf{98.8}&38.8/ 68.7 / 93.0&3.0 / 7.1 / 21.7&4.1 /15.5 / 21.4\\
				\hline
				\textbf{ParallelSearch+CRBNet}&\textbf{55.9} / \textbf{80.7} / \textbf{97.7}&\textbf{59.9} / \textcolor{red}{\textbf{83.8}} / \textbf{98.2}&\textbf{45.4} / \textbf{78.8} / \textbf{99.8}&\textbf{52.1} / \textbf{83.3} / \textbf{99.7}&\textcolor{red}{\textbf{62.9}} / 83.4 / \textcolor{red}{\textbf{98.8}}&\textcolor{red}{\textbf{59.9}} / \textbf{85.7} /\textcolor{red}{\textbf{ 98.4}}&\textcolor{red}{\textbf{51.5}} / \textbf{77.4} / \textbf{97.4}&\textcolor{red}{\textbf{10.7}} / \textcolor{red}{\textbf{23.3}} / \textcolor{red}{\textbf{39.0}}&\textbf{12.3} / \textcolor{red}{\textbf{29.8}} / \textcolor{red}{\textbf{39.3}}\\
				\hline
			\end{tabular}
		}
	\end{center}
	\label{table_robotcar}
\end{table*}
\newcommand{\tabincell}[2]{\begin{tabular}{@{}#1@{}}#2\end{tabular}} 
\begin{table}[h]
	\caption{Localization evaluation on Aachen Day-Night dataset. RT\_AP+CRBNet in the 7th line is a pure direct method with our CRBNet. ParallelSearch+CRBNet in the last line is ours of the new parallel search framework. The best results are in bold and the second-best are in red.}
	\begin{center}
		\begin{tabular}{ c | c | c } 
			\hline
			& day & night\\
			\cline{2-3}
			\multirow{2}{*}{\shortstack{distance[m]\\orient.[deg]} }& \multirow{2}{*}{\shortstack{.25/.50/5.0\\2/5/10} } & \multirow{2}{*}{\shortstack{0.5/1.0/5.0\\2/5/10} }\\
			& \multicolumn{1}{c|}{ } & \multicolumn{1}{c}{ }\\
			\hline
			
			
			
			
			AS \cite{sattler2016efficient} & 57.3/83.7/96.6 & 28.6/37.8/51.0\\
			CSL \cite{svarm2016city} & 52.3/80.0/94.3 & 39.6/40.8/56.1\\
			DenseVLAD \cite{torii201524} & 0.0/0.1/22.8 & 0.0/1.0/194\\
			NetVLAD \cite{arandjelovic2016netvlad} & 0.0/0.2/18.9 & 0.0/0.0/14.3\\
			FABMAP \cite{cummins2008fab}& 0.0/0.0/4.6 & 0.0/0.0/0.0\\
			NV+SP \cite{sarlin2019coarse} &\textbf{80.5}/87.4/94.2&\textbf{68.4}/77.6/88.8\\
			
			\textbf{RT\_AP+CRBNet} &75.0/\textcolor{red}{\textbf{92.4}}/\textbf{99.3} &
			54.1/73.5/88.8\\
			
			\textbf{IR+CRBNet} &73.9/91.4/96.7 & 61.2/\textcolor{red}{\textbf{81.6}}/\textcolor{red}{\textbf{89.8}}\\
			
			\textbf{RT+CRBNet} &74.0/91.0/99.0&51.0/67.3/81.6\\

			\textbf{ParallelSearch+CRBNet} &\textcolor{red}{\textbf{77.3}}/\textbf{94.3}/\textbf{99.3}&\textbf{68.4}/\textbf{86.7}/\textbf{96.9}\\
			\hline
		\end{tabular}
	\end{center}
	\label{table_aachen}
\end{table}

We compare RT\_AP+CRBNet with Active Search (AS) \cite{sattler2016efficient}, City-Scale Localization (CSL) \cite{svarm2016city}, DenseVLAD \cite{torii201524}, NetVLAD \cite{arandjelovic2016netvlad}, FAB-MAP \cite{cummins2008fab} and Hierarchical Localization (NV+SP) \cite{sarlin2019coarse}. The first two are direct methods, the following three are indirect methods, and the last one is serial indirect-direct method. 
AS searches matches not only from the direction of 2D-to-3D but also 3D-to-2D direction which allows to find more matches. CSL exploits ancillary gravitational information to help handle outliers. RT\_AP+CRBNet searches 2D-3D matches from 2D-to-3D direction and uses no additional information. NV+SP firstly uses NetVLAD to retrieve the top-k database images which are most similar to the given query image and then uses SuperPoint to search the nearest neighbor of the given query local feature among 3D points which can be seen by these retrieved images. All the results on RobotCar Seasons dataset are shown in Table \ref{table_robotcar} except for the last line. We see that RT\_AP+CRBNet has already performed better than all others except for NV+SP. Compared with NV+SP, there are 13 items, including "dawn", "dusk", "rain", that RT\_AP+CRBNet performs better, but in other 13 items, including "night" and "night-rain", NV+SP is better. So the performance of RT\_AP+CRBNet is on par with NV+SP.

RT\_AP+CRBNet just uses the supervised information of local descriptors to cluster 3D points and get nearest neighbor candidates of a given query local feature. Later, we will see if we use local and global information simultaneously to cluster 3D points and get nearest neighbor candidates of a given query local feature in the proposed parallel search framework, the performance is improved greatly as shown in the last line of Table \ref{table_robotcar}. 

The localization results on Aachen Day-Night dataset are shown in the first seven lines of Table \ref{table_aachen}, from which we can draw similar conclusions. More accurate results by our method of the parallel search framework will be given in the following section.
\subsection{Evaluation of the Proposed Parallel Search Framework}
In this subsection, the proposed parallel search framework for visual localization is evaluated. In the parallel search framework, we utilize random trees and NetVLAD to get the nearest neighbor candidates of a given query local feature and then perform a linear search in all candidates. The structure of random trees utilizes the local descriptor information while NetVLAD leverages global information. These two ways can complement each other and achieve higher accuracy.
We use NetVLAD to get top 20 frames.



The method of the proposed parallel search framework is denoted as ParallelSearch+CRBNet. The evaluation results of ParallelSearch + CRBNet on RobotCar Seasons dataset are shown in the last line in Table \ref{table_robotcar} including the comparisons with state-of-the-art methods. Compared with RT\_AP+CRBNet, ParallelSearch + CRBNet improves accuracies significantly. And ParallelSearch + CRBNet is also better than NV+SP in 19 items.
Evaluations on Aachen Day-Night dataset including comparisons are shown in Table \ref{table_aachen}. We see that ParallelSearch + CRBNet performs better than RT\_AP+CRBNet in all cases except that ParallelSearch + CRBNet is comparable with RT\_AP+CRBNet in one "day" item. 
Compared with NV+SP, ParallelSearch + CRBNet also performs better other than in a case of 'day' item. RT\_AP+CRBNet just uses the supervised information of local descriptors to construct random trees and then gets nearest neighbor candidates of query local features. NV+SP just utilizes global information to get nearest neighbor candidates of query local features. Different from these methods, our parallel search framework utilizes both local and global descriptor information to get nearest neighbor candidates of query local features and the comparison results show the advantages of the proposed parallel search strategy.

Experimental details of IR+CRBNet, and RT+CRBNet are given in Section 5 discussion.  

\section{Discussion} 

In this section, we report more experiments on Aachen Day-Night dataset \cite{sattler2018benchmarking}.



\subsection{The Influence of Patch Size}
\begin{figure}
	\centering
	\centerline{\includegraphics[width=8.0cm]{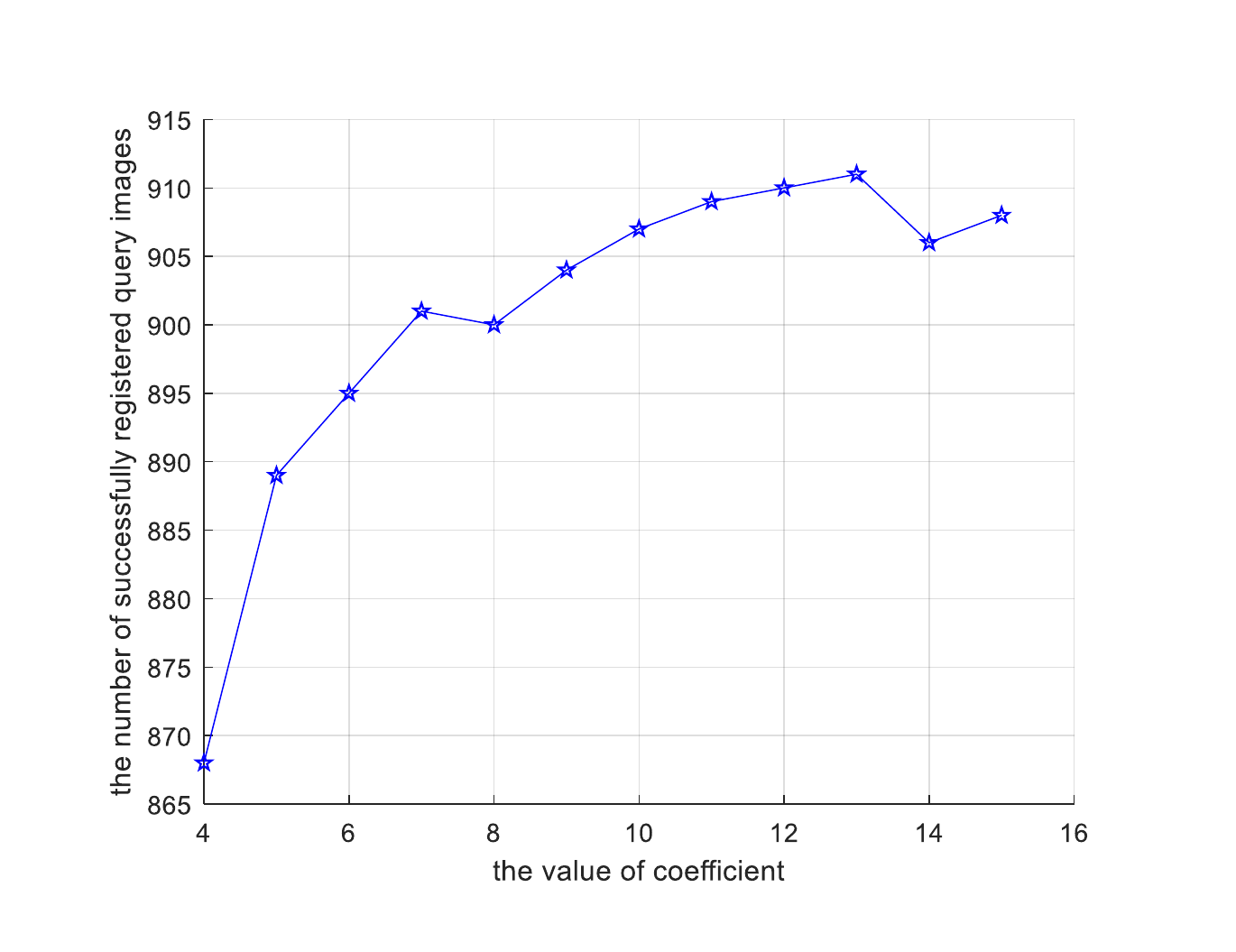}}
	\caption{The relationship between the used coefficient and the number of successfully registered query images in Aachen Day-night dataset \cite{sattler2018benchmarking}. }
	\label{coffi}
\end{figure}

The first step of our visual localization is to get our learning based local descriptors. The detailed process of getting our learning based local descriptors is as follows: 
\begin{itemize}	
	\item Use the same keypoint detector as for 3D reconstruction to get location, orientation, and scale.
	\item Sample a patch according to the product of the scale and a \textbf{coefficient} for each keypoint. 
	\item Rectify the patch using the orientation to eliminate a rotation between it and its matching patches.
	\item Resize the patch to 32\( \times \)32.
	\item Feed the patch into CRBNet and extract descriptors.	
\end{itemize}
The coefficient used in the above pipeline is very crucial. On one hand, the bigger the coefficient is, the more information the patch contains. On the other hand, if the coefficient is too large, we cannot crop a patch for the keypoint near to the boundary of a query image. So the coefficient directly affects the success rate of visual localization. We implement an experiment in Aachen Day-Night dataset \cite{sattler2018benchmarking} to explore the relationship between the coefficient and the number of successfully registered query images. The relationship between the coefficient and the number of successfully registered query images is shown in Fig. \ref{coffi}. 
As can be seen, when the coefficient is smaller than 13, the success rate increases with the coefficient increasing. But when the coefficient is bigger than 13, the curve begins to decrease which is in line with our analysis.

\subsection{Analysis of the Probability Model in the random trees}

The probability model with normal distribution in RT\_AP+CRBNet is substituted by the absolute value of the real-value descriptors. And this resulting method is denoted as RT+CRBNet. Its performance is also listed in Table \ref{table_aachen}. We can see that RT+ CRBNet is inferior to RT\_AP+CRBNet in all the cases. These results show that the probability model we used in random trees is very effective.


\subsection{Comparison of RT\_AP+CRBNet with IR+CRBNet}
We use NetVLAD to retrieve top 20 database images of a given query image and then search local correspondences by CRBNet. We name this serial indirect-direct method as IR+CRBNet. Its performance is also listed in Table \ref{table_aachen}. We observe that RT\_AP+CRBNet is superior to IR+CRBNet when localizing daytime query images. In terms of nighttime query images, the opposite is true. The reason may be that NetVLAD extracts global descriptors including high level semantic information which is more stable to large illumination change, so IR+CRBNet can retrieve KNNs of the given query image correctly by NetVLAD even illumination varies greatly. However, RT\_AP+CRBNet is based on local descriptors with lower level semantic information which is not as robust to illumination as global descriptors. But in daytime, when the viewpoint varies large, influence to local descriptors is slighter than to global descriptors. This is why RT\_AP+CRBNet achieves higher accuracy in daytime. This confirms our idea that local and global descriptors are complementary when finding nearest neighbor candidates of a given query local feature also as shown in the last line by ParallelSearch+CRBNet in this table.



\section{Conclusion and Future Work}

In this paper, we present a novel framework of fusing local and global feature information to select nearest neighbor candidates of a given query local feature for visual localization.
On one hand, we use random trees structured by the binary CRBNet to index 3D points in databases. Then a probability model for our learning based local descriptors is given to get prior leaf nodes. On the other hand, we extract global descriptors for the whole images and perform a global image retrieval to obtain prior frames. The 3D points in prior leaf nodes and visible in prior frames are as nearest neighbor candidates of the given query local feature. Finally, local 2D-3D correspondences are established by the real-valued form of CRBNet and 6DoF pose of a given query image is calculated by a PnP-RANSAC scheme.
Experiments show that the proposed localization method can significantly improve the robustness and accuracy compared with the ones which get nearest neighbor candidates of a given query local feature just based on either local or global descriptors.
In future work, we will try to learn more discriminative descriptors and then combine with our parallel framework to improve the accuracy of visual localization further.

\section*{Acknowledgment}
This work was supported by National Natural Science
Foundation of China under Grant Nos. 61836015, 61421004 and supported by the Beijing Advanced Discipline Fund under Grant No. 115200S001.


\bibliography{egbib}

\end{document}